\documentclass[journal,transmag]{IEEEtran}

\usepackage{times}
\usepackage{epsfig}
\usepackage{graphicx, subfigure}
\usepackage{lscape}

\usepackage{graphicx}
\usepackage{amsmath}
\usepackage{amssymb}
\usepackage{amsthm}
\usepackage{cite}
\usepackage{url}
\usepackage{color}
\usepackage{tikz}
\usepackage{threeparttable}
\usepackage{multirow}
\usepackage{multicol}

\usepackage{makeidx, tabularx, graphicx, setspace, amsmath, longtable, multirow, float, theorem, times, booktabs, siunitx, lineno}

\usepackage{lipsum}
\usepackage{cuted}

\usepackage{mathtools,amssymb,lipsum}

\usepackage{cuted}
\definecolor{Red}{rgb}{1, 0.2, 0.2}

\newcommand{\zzz}[1]{\textcolor{black}{#1}}
\newcommand{\ling}[1]{\textcolor{black}{#1}}

\ifCLASSINFOpdf

\else

\fi

\newcommand\copyrighttext{%
\centering
  \footnotesize Copyright (c) 2019 IEEE. Personal use of this material is permitted. However, permission to use this material for any other purposes must be obtained from the IEEE by sending a request to pubs-permissions@ieee.org.}
\newcommand\copyrightnotice{%
\begin{tikzpicture}[remember picture,overlay]
\node[anchor=south,yshift=10pt] at (current page.south) {\parbox{\dimexpr\textwidth-\fboxsep-\fboxrule\relax}{\copyrighttext}};
\end{tikzpicture}%
}

\begin{document}

%
% paper title
% can use linebreaks \\ within to get better formatting as desired
% Do not put math or special symbols in the title.
\title{Spatio-Temporal Convolutional LSTMs for Tumor Growth Prediction by Learning 4D Longitudinal Patient Data
\\
%\xxx{LOOK FOR RED TEXT FROM MS - that one needs our discussion/answers}
}

% author names and affiliations
% transmag papers use the long conference author name format.

\author{\IEEEauthorblockN{Ling Zhang\IEEEauthorrefmark{},
Le Lu\IEEEauthorrefmark{},~\IEEEmembership{Senior Member,~IEEE},
\zzz{Xiaosong Wang}\IEEEauthorrefmark{},
Robert M. Zhu\IEEEauthorrefmark{},
Mohammadhadi Bagheri\IEEEauthorrefmark{},\\
Ronald M. Summers\IEEEauthorrefmark{}, and 
Jianhua Yao\IEEEauthorrefmark{}}

\thanks{Manuscript received XX YY, 2019. This work was supported by the Intramural Research Program at the NIH Clinical Center. The authors thank Nvidia for the TITAN X Pascal GPUs donation. \emph{Asterisk indicates corresponding authors.}}
\thanks{*L. Zhang is with \zzz{PAII Inc.}, Bethesda, MD \zzz{20817}, USA (e-mail: zhangling0722@163.com).}
\thanks{L. Lu is with PAII Inc., Bethesda, MD 20817, and Johns Hopkins University, Baltimore, MD 21218, USA.}
\thanks{\zzz{X. Wang is with Nvidia Corporation, Bethesda, MD 20814, USA.}}
\thanks{R. M. Zhu, M. Bagheri, and R. M. Summers are with the Imaging Biomarkers and Computer-Aided Diagnosis Laboratory and Clinical Image Processing Service, Radiology and Imaging Sciences Department, National Institutes of Health Clinical Center, Bethesda, MD 20892, USA.}
\thanks{*J. Yao is with the Tencent Holdings Limited, Shenzhen 518057, China (e-mail: jianhua\_yao@yahoo.com).}
\thanks{The majority part of work was performed when L. Zhang was with National Institutes of Health \zzz{and Nvidia Corporation}, L. Lu and J. Yao were with National Institutes of Health Clinical Center.}
}

% The paper headers
\markboth{}%
{Zhang \MakeLowercase{\textit{et al.}}: Spatio-Temporal Convolutional LSTMs of Learning 4D Longitudinal Data for Tumor Growth Prediction}

\maketitle

\begin{abstract}
Prognostic tumor growth modeling via volumetric medical imaging observations can potentially lead to better outcomes of tumor treatment management and surgical planning. %Traditionally, this problem is tackled through mathematical modeling and evaluated using relatively small patient datasets. 
Recent advances of convolutional networks (ConvNets) have demonstrated higher accuracy than traditional mathematical models can be achieved in predicting future tumor volumes. This indicates that deep learning based data-driven techniques may have great potentials on addressing such problem. %The current state-of-the-art work models the cell invasion and mass-effect of tumor growth by training separate ConvNets on 2D image patches to reveal both tumor's static and dynamic information derived from medical imaging data. 
However, \zzz{current 2D image patch based} modeling approaches can not make full use of the spatio-temporal imaging context of the tumor's longitudinal 4D (3D + time) patient data. Moreover, they are incapable to predict clinically-relevant tumor properties, other than the tumor volumes. 
In this paper, we exploit to formulate the tumor growth process through convolutional \zzz{Long Short-Term Memory} (ConvLSTM) that extract tumor's static imaging appearances and simultaneously capture its temporal dynamic changes within a single network. We extend ConvLSTM into the spatio-temporal domain (ST-ConvLSTM) by jointly learning the inter-slice 3D contexts and the longitudinal or temporal dynamics from multiple patient studies. Our approach can incorporate other non-imaging patient information in an end-to-end trainable manner. Experiments are conducted on the largest 4D longitudinal tumor dataset of 33 patients to date. Results validate that the proposed ST-ConvLSTM model produces a Dice score of 83.2\%$\pm$5.1\% and a RVD of 11.2\%$\pm$10.8\%, both statistically significantly outperforming ($p<$0.05) other compared methods of traditional linear model, ConvLSTM, and generative adversarial network (GAN) under the metric of predicting future tumor volumes. Additionally, our new method enables the prediction of both cell density and CT intensity numbers. %In addition, our model could predict tumor growth at later time point in promising accuracy with RVD of 37.2\%$\pm$42.5\%. %The proposed ST-ConvLSTM can be potentially applied to predict any 3D/4D medical imaging at future time points.
\zzz{Last, we demonstrate the generalizability of ST-ConvLSTM by employing it in 4D medical image segmentation task, which achieves an averaged Dice score of 86.3\%$\pm$1.2\% for left-ventricle segmentation in 4D ultrasound with 3 seconds per patient case.}
\end{abstract}

% Note that keywords are not normally used for peerreview papers.
\begin{IEEEkeywords}
Tumor growth prediction, Deep learning, Convolutional LSTM, Spatio-temporal Longitudinal Study, \zzz{4D Medical Imaging}.
\end{IEEEkeywords}

\copyrightnotice

% make the title area
\maketitle

\IEEEdisplaynontitleabstractindextext

\IEEEpeerreviewmaketitle

%serial long-term (e.g., $\geq 6$ months) 
% \xxx{xxxxxxxxxxxxxxxx--to-here}

\section{Introduction}

Tumor growth modeling using medical images of longitudinal studies is a challenging yet important problem in precision and predictive medicine, because it may potentially lead to better tumor treatment management and surgical planning for patients. \ling{For example, treatments of pancreatic neuroendocrine tumor (PanNET or PNET) include active surveillance, surgical intervention, and medical treatment. Active surveillance is undertaken if a PanNET does not reach 3 cm in diameter or a tumor-doubling time $<$500 days; otherwise the corresponding PanNET should be resected due to the high risk of metastatic disease \cite{keutgen2016evaluation}. Medical treatment (e.g., everolimus) is for the intermediate-grade (PanNETs with radiologic documents of progression within the previous 12 months), advanced or metastatic disease \cite{yao2011everolimus}. Therefore the patient-specific prediction of PanNET's growth pattern at earlier stages is highly desirable, since it will assist decision making on different treatment strategies to better manage the undergoing treatment or surgical planning.} 

Conventionally, this task has been well exploited through complex and sophisticated mathematical modeling \cite{swanson2000quantitative,clatz2005realistic,hogea2008image,menze2011generative,liu2014patient,wong2017pancreatic,roque2018dce}, which accounts for both cell invasion and mass-effect using reaction-diffusion equations and bio-mechanical models. From there the actual tumor growth can be predicted by personalizing the established model based on clinical imaging derived tumor physiological parameters, such as morphology, metabolic rate, and cell density. While these methods yield informative results, most of them have not been able to utilize the underlying statistical distributions of tumor growth patterns in the studied patient population. The number of mathematical model parameters is often very limited (e.g., 5 in \cite{wong2017pancreatic}), which might not be sufficient to model the inherent complexities of the growing tumors. 

Furthermore, two alternative approaches have been proposed to predict tumor growth. 1) Assuming that the future tumor growth pattern follows its past trend, optical flow computing can be used to estimate previous voxel-level tumor motions, and subsequently, to predict the future deformation field via an autoregressive model \cite{weizman2012prediction}. Therefore the entire future brain MR scan can be generated but the resulting tumor volume still needs to be measured manually. \zzz{For slow- and fast-growing brain tumors, the method achieves 13.7\% and 34.2\% volumetric estimation errors, respectively.} \zzz{However this approach does not involve the tumor growth pattern in population trend, and} may over-simplify the essential challenge because it only infers the future tumor imaging under a linear way. 2) To address this issue, machine learning principle is a potential solution to incorporate the population trend into tumor growth modeling. The pioneer study \cite{morris2006learning} attempts to model the glioma growth patterns as a pixel classification problem where traditional machine learning pipeline of hand-crafted feature extraction and selection and classifier training is applied. \zzz{Although only moderate levels of accuracy (where both precision and recall values are 59.8\% \cite{morris2006learning}) has been achieved, this data-driven statistical learning approach has shown its potential to tackle the highly changeling task of glioma (as a fast-growing tumor) growth prediction. Nevertheless the hand-crafted imaging features could be compromised by the limited understanding of tumor growth process, and may not generalize well for other tumors.}
%conventional statistical techniques used in this study \zzz{may} not \zzz{be} capable to achieve satisfying prediction results, on the complex task of tumor growth prediction.
%\zzz{For example, the prediction of glioma (a fast-growing tumor) growth has just moderate accuracy, with both precision and recall are 59.8\% \cite{morris2006learning}.}

Recently, statistical and deep learning framework \cite{zhang2017personalized} and two-stream convolutional neural networks (ConvNets) \cite{zhang2018convolutional} have shown more compelling and improved performance than the mathematical modeling approach \cite{wong2017pancreatic} using the same pancreatic tumor dataset. More importantly, the later study \cite{zhang2018convolutional} demonstrate the effectiveness of deep ConvNets in characterizing two fundamental processes of both cell invasion and mass-effect of tumor growth. 

From \cite{zhang2017personalized}, image patch based ConvNets extract deep image features that are late-fused with clinical factors, followed by a support vector machine (SVM) classifier using all features. Such a separated process may not fully exploit the inherent correlations between the deep image features and clinical factors. The two-stream ConvNet architecture \cite{zhang2018convolutional} treats the prediction as a local patch-based classification task, which does not consider the global information of the tumor structure and its surrounding spatio-temporal context. Both methods make predictions based on 2D image slices whereas the tumor growth modeling is in fact a 4D (3D+time) problem. Additionally, \cite{zhang2017personalized,zhang2018convolutional} cannot predict other clinically relevant properties, such as tumor cell density and radiodensity in Hounsfield units (HU). Last, due to the difficulties in collecting the longitudinal tumor data and the complexities of data preprocessing, 
both studies are only conducted and evaluated using a relatively small dataset consisting of ten patients. 

In this paper, we propose a novel deep learning approach that incorporates both 3D spatial and temporal image properties and clinical information into one single deep neural network. %, enable the prediction of tumor cell density and radiodensity, and evaluate the presented approach on a much larger dataset of 33 patients. 
Our main contributions are summarized as follows. 
(1) A novel spatio-temporal Convolutional Long Short-Term Memory (ST-ConvLSTM) network is proposed to jointly learn the intra-slice spatial structures, the inter-slice correlations in 3D contexts, and the temporal dynamics in time sequences.
(2) Compared to previous machine (deep) learning based methods \cite{morris2006learning,zhang2017personalized,zhang2018convolutional} that utilize 2D image patches and predict the future tumor volume only, our new model is holistic image-based and enables the predictions of future tumor imaging properties, i.e., future cell density and CT intensity numbers for relevant clinical diagnosis. 
(3) Other clinical information, such as time intervals can be fully integrated into our end-to-end trainable deep learning framework. %, allowing prediction for the later time points. 
(4) To the best of our knowledge, we construct the largest longitudinal pancreatic neuroendocrine tumor \zzz{(which is a relatively slow-growing tumor)} growth database to date (33 patients \zzz{with serial CT imaging added}), \zzz{enriching our previous dataset by more than 3 times \cite{zhang2018convolutional,zhang2017personalized}}. %, enabling the first statistically sound study in the field of tumor growth prediction. 
\zzz{(5) We demonstrate the effectiveness and high efficiency of employing 4D ST-ConvLSTM for a 3D+time left-ventricle ultrasound image segmentation task. Only a small subset of sparsely-annotated 3D ultrasound volumes per time sequence are required by ST-ConvLSTM.}

\section{Related Work}
%\subsection{Future Image Frame Prediction Using Deep Learning}
%since it is closely-related to unsupervised feature learning and could enable intelligent agents to react to the environments. 
In recent computer vision developments, the task of future image frame prediction (i.e., predicting a visual pattern of RGB raw pixels given a short video sequence) has attracted great research interests \cite{villegas2017decomposing,lu2017flexible,kalchbrenner2017video,liang2017dual,wang2017predrnn,liu2018future}. It is closely related to unsupervised feature learning and can enable intelligent agents to react to the environments. Table \ref{literature} briefly summarizes recent representative deep learning based approaches to tackle this problem. There are mainly four key technique components being exploited: convolutional LSTMs (ConvLSTM), generative adversarial network (GAN), encoder-decoder network, and motion (mostly optical flow) cues. 

\begin{table}[!t] 
\scriptsize
\centering
\caption{Deep Learning Based Future Image Frame Prediction Methods and Their Key Techniques. ConvLSTM: Convolutional Long Short-Term Memory; GAN: Generative Adversarial Network.
}
\label{literature}
\begin{tabular}{p{2cm}|p{1.2cm}|p{0.9cm}|p{1.8cm}|p{0.9cm}}
\hline
 & ConvLSTM & GAN & Encoder-Decoder & Motion\\
\hline
LSTM \cite{srivastava2015unsupervised} & - & - & - & \\
\hline
ConvLSTM \cite{xingjian2015convolutional} & $\surd$ & - & - & - \\
\hline
BeyondMSE \cite{mathieu2015deep} & - & $\surd$ & - & - \\
\hline
Autoencoder \cite{patraucean2016spatio} & $\surd$ & - & $\surd$ & $\surd$\\
\hline
CDNA\cite{finn2016unsupervised} & $\surd$ & - & - & $\surd$ \\
\hline
MCNet \cite{villegas2017decomposing} & $\surd$ & - & $\surd$ & $\surd$ \\
\hline
PredNet \cite{lotter2016deep} & $\surd$ & - & - & - \\
\hline
STNet \cite{lu2017flexible} & $\surd$ & $\surd$ & $\surd$ & $\surd$ \\
\hline
VPN \cite{kalchbrenner2017video} & $\surd$ & - & $\surd$ & - \\
\hline
Hierarchical \cite{villegas2017learning} & - & - & $\surd$ & - \\
\hline
S2S-GAN \cite{luc2017predicting} & - & $\surd$ & - & - \\
\hline
DVF \cite{liu2017video} & - & - & $\surd$ & $\surd$ \\
\hline
DM-GAN \cite{liang2017dual} & $\surd$ & $\surd$ & $\surd$ & $\surd$ \\
\hline
PredRNN \cite{wang2017predrnn} & $\surd$ & - & - & - \\
\hline
SNCCL \cite{bhattacharjee2017temporal} & - & $\surd$ & - & - \\
\hline
Two-stream \cite{jin2017predicting} & - & - & $\surd$ & $\surd$ \\
\hline
Spatial-motion \cite{liu2018future} & $\surd$ & $\surd$ & $\surd$ & $\surd$ \\
\hline
\end{tabular}
\end{table}

%-------------
   \begin{figure*}[!t]
   \begin{center}
   \begin{tabular}{c}
   \includegraphics[width=15cm]{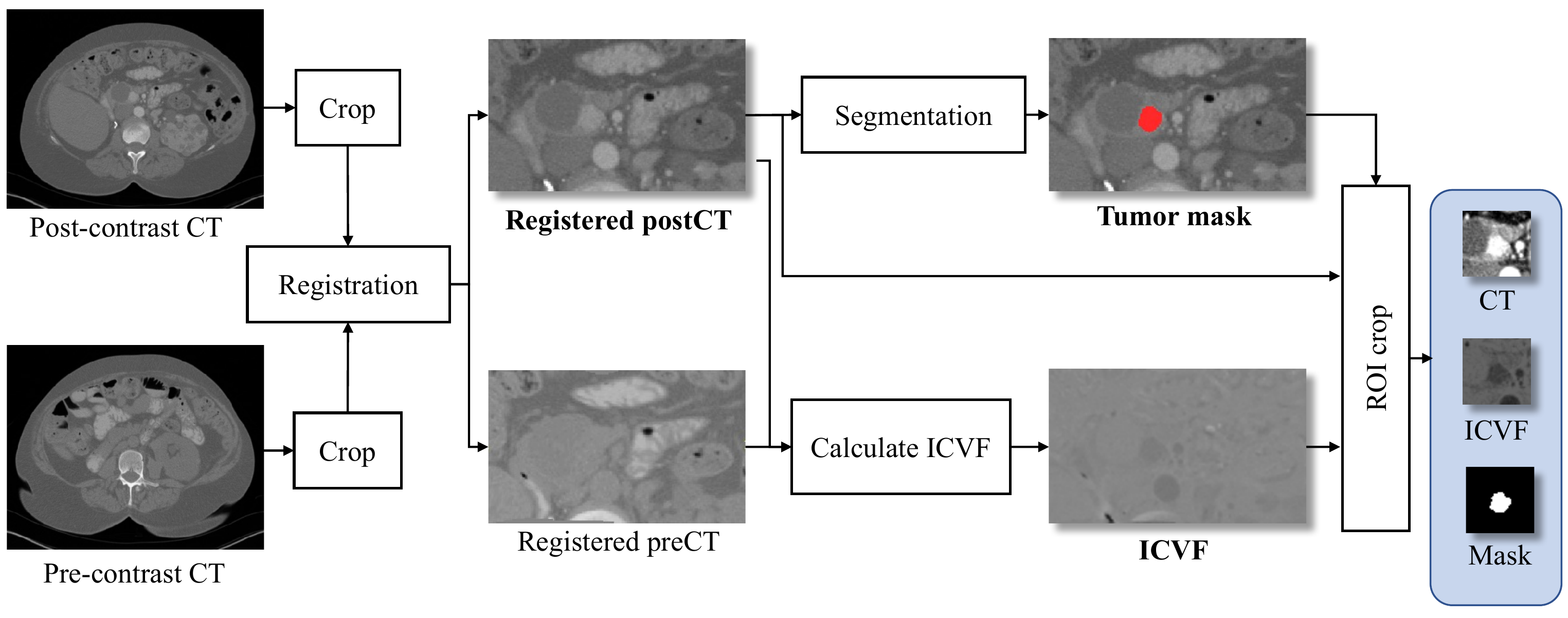}
   \end{tabular}
   \end{center}
   \caption
   { \label{fig_data} 
Image processing pipeline of constructing the tumor dataset for one time point. 
}  
   \end{figure*} 
%-------------

LSTM \cite{hochreiter1997long} is designed for the next time-step status prediction in a temporal sequence, and can be naturally extended to predict the consequent frames from previous ones in a video \cite{srivastava2015unsupervised}. Next, ConvLSTM \cite{xingjian2015convolutional} is proposed to preserve the spatial structure in both the input-to-state and state-to-state transitions. Subsequently, ConvLSTM becomes the backbone model of several video prediction approaches \cite{patraucean2016spatio,finn2016unsupervised,villegas2017decomposing,lotter2016deep,lu2017flexible,kalchbrenner2017video,liang2017dual,wang2017predrnn,liu2018future}, where each work is enhanced with additional improvements. For example, 1) optical flow is introduced in an encoder-ConvLSTM-decoder framework \cite{patraucean2016spatio} to explicitly model the temporal dynamics; %The idea is to first use ConvLSTM to predict the optical flow (deformation map), which is then used to transform the current frame to generate next frame;
2) ConvLSTM is reformulated to predict motions from the current pixels to the next pixels \cite{finn2016unsupervised} with the goal of alleviating the blurry prediction images; 3) ConvLSTM is integrated in encoder-decoder networks to estimate the discrete joint distributions of the RGB pixels which archived the highest accuracy on the moving digits dataset \cite{kalchbrenner2017video}; 
%A combination of ConvLSTM and optical flow in a generative adversarial network (GAN) learning framework is proposed in \cite{liang2017dual}, where state-of-the-art performance is archived on vehicle and pedestrian datasets.
4) additionally, a new spatiotemporal LSTM unit \cite{wang2017predrnn} is designed to memorize both temporal and spatial representations, thus obtaining better performances than the conventional LSTM. 

In addition to ConvLSTM, ConvNets integrated with GAN \cite{mathieu2015deep,luc2017predicting,bhattacharjee2017temporal} based image generators represent the other thread of promising solutions, especially effective on sharpening blurry predictions. %caused by the $\ell_{2}$ loss function. 
Encoder-decoder networks \cite{patraucean2016spatio,villegas2017decomposing,lu2017flexible,kalchbrenner2017video,villegas2017learning,liu2017video,liang2017dual} commonly serve as backbone deep learning architectures to achieve the image-to-image prediction that typically contain multiple convolutional layers for subsampling and several deconvolutional layers for upsampling. Comprehensive discussions of the above techniques are given in \cite{lu2017flexible,liang2017dual,liu2018future}, where state-of-the-art quantitative performances are presented using video, vehicle and pedestrian datasets.

%\subsection{3D Medical Image Segmentation Using Deep Learning}

ConvLSTM has also been employed for 3D medical image segmentation, and is an effective way of treating the 3D volume as a sequence of 2D consecutive slices \cite{chen2016combining,cai2017improving,tseng2017joint}. Compared to the 2D ConvNets-based segmentation, ConvLSTM tends to be more robust and consistent inter-slice wise since 3D contextual information is memorized and propagated in the $z$-direction. 

Beyond the problems of 3D medical image segmentation (directly on 3D volumetric data scans) and natural video prediction (using 2D image+time sequences), tumor growth prediction is processed on 4D longitudinal volumetric patient imaging scans. %, which include the spatial structure in the $xy$ plane, the sequential dependency in the $z$-direction, and the temporal dynamics in the time sequence. 
Desirable prediction models should not only recall the temporal evolution trend, but also keep consistent with the tumor's 3D spatial contexts. Motivated by this assumption, we propose a novel ST-ConvLSTM network to explicitly capture their dependencies among 2D image slices, through the recurrent analysis over spatial and temporal dimensions concurrently. %Motivated by this, we propose a novel spatio-temporal ConvLSTM (ST-ConvLSTM) network which jointly learns the intra-slice spatial structure, the inter-slice correlations in 3D contexts, and the temporal dynamics in time sequence. %There are few LSTM-based models that learn different spatio-temporal information, such as video prediction \cite{wang2017predrnn} and 3D action recognition \cite{liu2018skeleton}.
\zzz{An alternative way of extending ConvLSTM to 4D image is based on 3D ConvLSTM model \cite{yang2017deep}. However, given the computational complexity of both 3D convolution and LSTM, such a model is hard to train and get converged. Furthermore, due to the large GPU memory consumption, its input size is limited which potentially affects its performance.} 

\zzz{4D medical image segmentation, such as the segmentation of 3D+time ultrasound volumes \cite{bernard2015standardized}, is another application scenario of our ST-ConvLSTM model. Currently, 2D or 3D ConvNets are the main deep neural network models to solve 4D segmentation problem. Although well-designed 2D/3D ConvNets could produce promising accuracy, the 4th temporal dimension contains the time-consistency constraints and can potentially improve the 4D segmentation accuracy. However direct usage of 4D ConvNets for segmentation is extremely slow and less practical, mainly because of the large computational complexity and the lack of 4D labels (e.g., the manual segmentation annotations for all 3D image volumes per sequence).}

%-------------
   \begin{figure*}[!t]
   \begin{center}
   \begin{tabular}{c}
   \includegraphics[width=17cm]{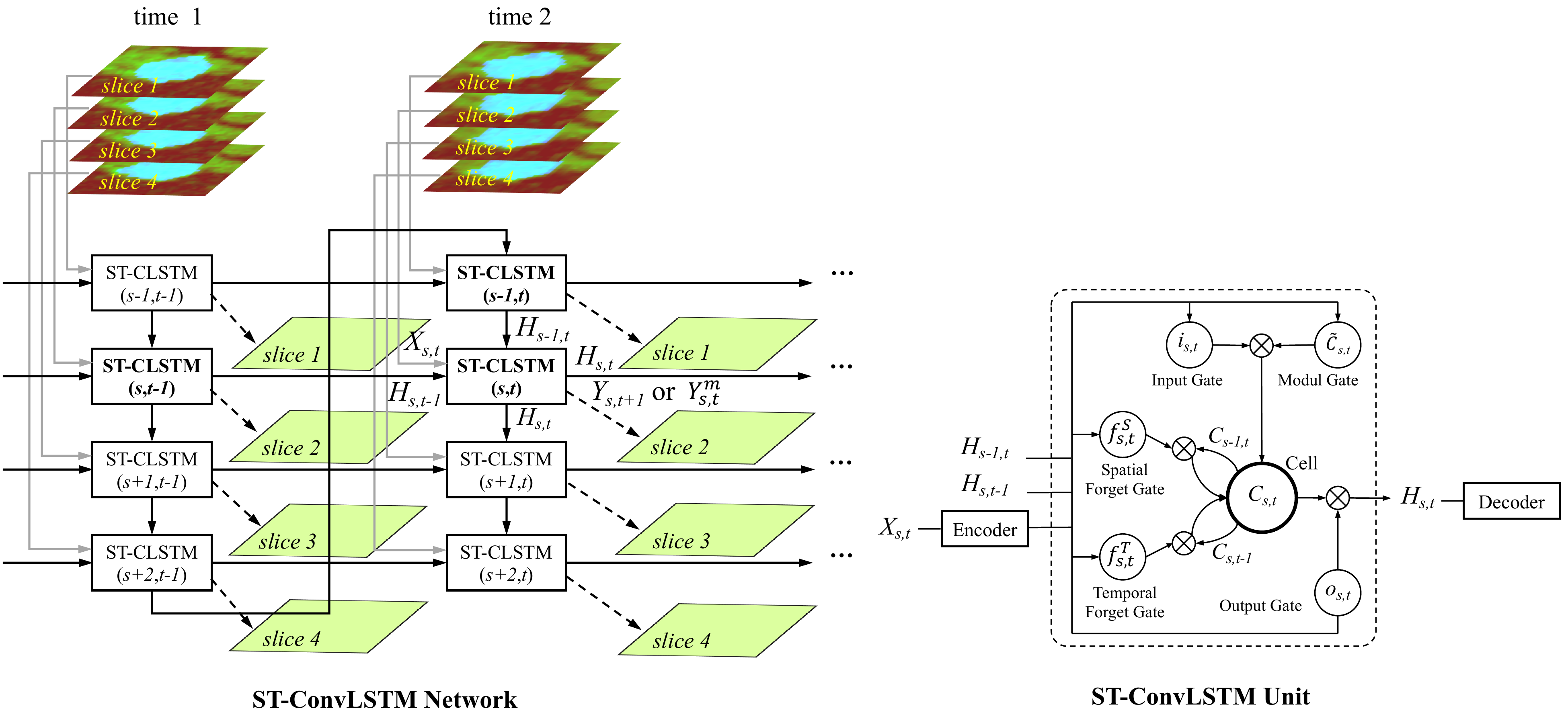}
   \end{tabular}
   \end{center}
   \caption
   { \label{fig_STCLSTM_cell} 
{\bf Left:} The proposed Spatio-Temporal Convolutional LSTM (ST-ConvLSTM, or ST-CLSTM) network for learning of 4D \zzz{medical imaging representations} to predict tumor growth \zzz{or segment object}. In this example, 2 time points (each with 4 spatially adjacent image slices and each slice is a 3-channel color image) are shown. %Given the tumor imaging data from time 1 and time 2, our model is trained to predict the tumor imaging at time 3. 
\zzz{This network model can be either used to predict tumor growth in 4D longitudinal data (i.e., to generate a future slice $Y_{s,t+1}$) given the input $X_{s,t}$; or segment objects in 3D+time images (i.e., to compute the current segmentation mask frame $Y_{s,t}$ from an input ultrasound image $X_{s,t}$ in Sec. \ref{sec:ultrasound}).} {\bf Right:} The ST-ConvLSTM unit. \zzz{The encoder-decoder architecture is depicted in Fig. \ref{fig_architecture_feature}.}
}  
   \end{figure*} 
%-------------

\section{Methods}

\subsection{Construction of 4D Longitudinal Tumor Dataset}
\label{dataconstruction}

Our 4D longitudinal tumor imaging data set used in this study consists of dual-phase contrast-enhanced CT volumes at three time points for each patient. As shown in Fig. \ref{fig_data}, for \zzz{each pair of} pre- and post-contrast (arterial phase) \zzz{3D CT volumes at the same time point, their organ (e.g., pancreas) regions} are first \zzz{roughly cropped and} registered \zzz{to post-contrast CT} using the ITK\footnote[1]{https://itk.org/} implementation of mutual information based B-spline registration \cite{rueckert1999nonrigid}. The segmentation is performed manually by a medical trainee using ITK-SNAP \cite{neuroimage2006}\footnote[2]{http://www.itksnap.org/} on the post-contrast CT (as those tumors can be better evaluated in the arterial phase), under supervision of an experienced radiologist. Three image feature channels are derived: 1) intracellular volume fraction (ICVF) images representing the cell density that is normalized between [0 100] (more details about ICVF calculation can be referred to \cite{liu2014patient}); 2) post-contrast CT images in soft-tissue window [-100, 200HU] and linearly transformed to [0 255]; 3) binary tumor segmentation mask (0 or 255). A sequence of image patches of 32$\times$32 pixels\footnote[3]{Most pancreatic tumors in our dataset are $<$3 cm ($\approx$30 pixels) in diameter.} centered on the 3D tumor centroid is cropped to cover the entire tumor. The cropping is repeated for the three ICVF-CT-Mask channels (right panel in Fig. \ref{fig_data}) and forms an RGB image as illustrated in Fig. \ref{fig_STCLSTM_cell}. The dataset is prepared for every tumor volume at each time point, and imaging volumes at different times are aligned using the segmented 3D tumor centroids, to build the spatio-temporal sequence data set for training and testing. \zzz{We acknowledge that there might be some bias of using simple tumor centroids for the longitudinal alignment, but this is a relatively (more) reliable approach compared to the image appearance based registration methods, based on our preliminary experiment and past studies (e.g., \cite{wong2017pancreatic,zhang2017personalized,zhang2018convolutional})}.

\subsection{Spatio-Temporal Convolutional LSTM} 

\subsubsection{Convolutional LSTM}

LSTM \cite{srivastava2015unsupervised} operates on temporal sequences of 1D vectors, and can reconstruct the input sequences or predict the future sequences. A LSTM unit contains a memory cell $C_{t}$, an input gate $i_{t}$, a forget gate $f_{t}$, an output gate $o_{t}$, and an output state $H_{t}$. Compared with the conventional LSTM, ConvLSTM is capable of modeling 2D spatio-temporal image sequences by explicitly encoding their 2D spatial structures (replacing LSTM's fully connected transformations with spatial local convolutions in ConvLSTM) into the temporal domain \cite{xingjian2015convolutional,cai2017improving}. The main equations of ConvLSTM are as follows:

\begin{small}
\begin{align}
\begin{split}
%\hspace{-6cm}
f_{t} &= \sigma(W_{xf}*X_{t} + W_{hf}*H_{t-1} + b_{f}) \\ 
i_{t} &= \sigma(W_{xi}*X_{t} + W_{hi}*H_{t-1} + b_{i}) \\ 
\widetilde{C}_{t} &= {\rm tanh}(W_{x\widetilde{C}}*X_{t} + W_{h\widetilde{C}}*H_{t-1} + b_{\widetilde{C}}) \\
o_{t} &= \sigma(W_{xo}*X_{t} + W_{ho}*H_{t-1} + b_{o}) \\
C_{t} &= f_{t}\odot C_{t-1} + i_{t}\odot \widetilde{C}_{t} \\
H_{t} &= o_{t}\odot {\rm tanh}(C_{t})
\end{split}
\end{align}
\end{small}

where $\sigma$ and $\rm tanh$ are the sigmoid and hyperbolic tangent non-linearities, $*$ is the convolution operator, and $\odot$ is the Hadamard product. The input $X_{t}$, cell $C_{t}$, hidden states $H_{t}$,  \zzz{forget gate} $f_{t}$, \zzz{input gate} $i_{t}$, \zzz{input-modulation gate} $\widetilde{C}_{t}$, and \zzz{output gate} $o_{t}$ are all 3D tensors with the dimension of $M\times N\times F$ (rows, columns, feature maps). The memory cell $C_{t}$ is the key module, which acts as an accumulator of the state information controlled by the gates.

\subsubsection{ST-ConvLSTM Network and Unit}
Given the ICVF-CT-Mask three-channel {\bf input} maps at time 1 and time 2 \zzz{(as $X_{t} = \{X^{i}_{t},X^{c}_{t},X^{m}_{t}\}, t\in\{1,2\}$, respectively)}, the aim is to predict the \zzz{{\bf output}} ICVF-CT-Mask maps at time 3 \zzz{(as $Y_{t} = \{Y^{i}_{t},Y^{c}_{t},Y^{m}_{t}\}, t=3$)}, shown in Fig. \ref{fig_STCLSTM_cell}. Directly using ConvLSTM over temporal domain could discover the tumor 2D dynamics for its growth prediction. Furthermore, the spatial consistency in the 3D volume data and its form of sequential nature of 2D image slices make it possible to extend ConvLSTM to the 3D spatial domain. 

%-------------
   \begin{figure*}[!t]
   \begin{center}
   \begin{tabular}{c}
   \includegraphics[width=15cm]{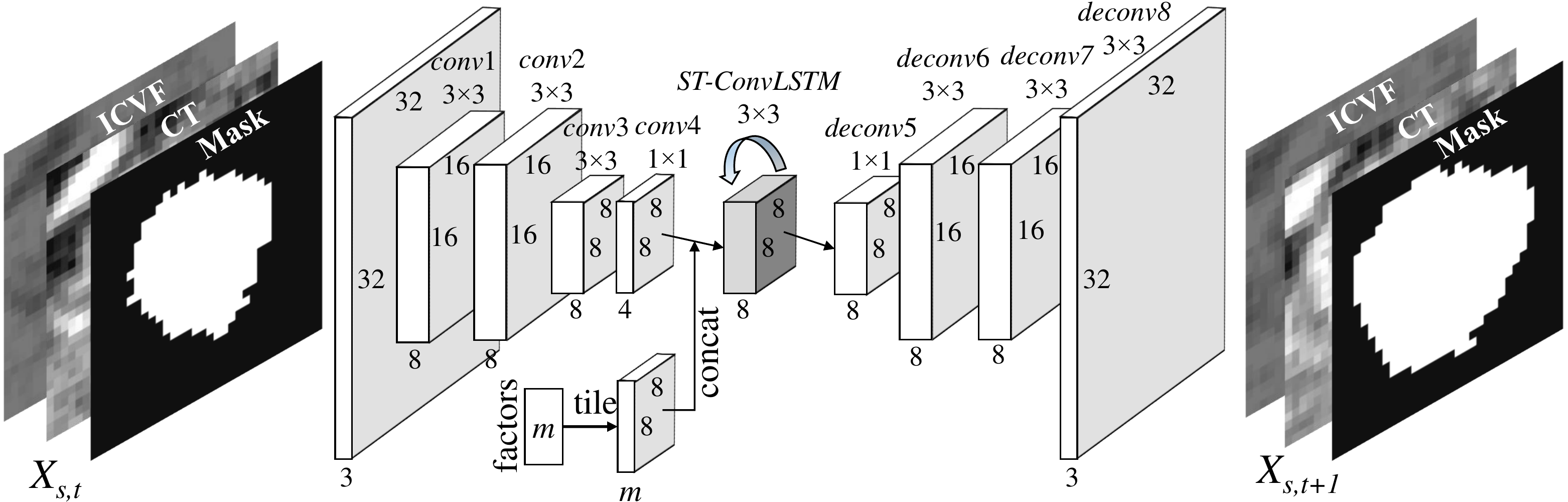}
   \end{tabular}
   \end{center}
\caption{The end-to-end network architecture of our proposed encoder-ST-ConvLSTM-decoder for tumor growth prediction. \zzz{For 4D segmentation task, the input is replaced with the raw (e.g., ultrasound) image, the output is its mask, no ``factor" branch for other clinical properties, and network model channels are set to 1-8-16-32-64-64-64-32-16-1.}}
\label{fig_architecture_feature} 
   \end{figure*} 
%-------------

Instead of simply concatenating the 2D CT slices, in order to learn simultaneously both the spatial consistency patterns among successive image slices and the temporal dynamics across different time points, we propose a new Spatio-Temporal Convolutional LSTM (ST-ConvLSTM) network as illustrated in Fig. \ref{fig_STCLSTM_cell} (left panel). In this network, each ST-ConvLSTM unit takes input from one image slice at one time point in the 4D space, and receives the hidden states from both the horizontal (the same slice locations at previous time) and vertical directions (previous adjacent slice at the current time). For example, the unit $(s,t)$ in Fig. \ref{fig_STCLSTM_cell} (left panel) corresponds to the $s^{th}$ slice at time $t$, and receives the hidden states $H_{s,t-1}$ from unit $(s,t-1)$ and $H_{s-1,t}$ from unit $(s-1,t)$. Along with the current input image slice $X_{s,t} \zzz{= \{X^{i}_{s,t},X^{c}_{s,t},X^{m}_{s,t}\} }$, the ST-CLSTM unit $(s,t)$ can \zzz{\textit{predict}} the future slice $Y_{s,t+1} \zzz{= \{Y^{i}_{s,t+1},Y^{c}_{s,t+1},Y^{m}_{s,t+1}\} } $, and generate its hidden state $H_{s,t}$. \zzz{For the 4D ultrasound image segmentation task, the goal is from any current input raw image slice (e.g., $X_{s,t}$) to generate its \textit{output segmentation} mask $Y^{m}_{s,t}$.} In each ST-CLSTM unit ({\bf right} in Fig. \ref{fig_STCLSTM_cell}), since there are two different candidates generated from the spatial and temporal domains, respectively, two forget gates $f_{s,t}^{S}$ and $f_{s,t}^{T}$ are equipped for adding them to update the unit state. The activations of a ST-ConvLSTM at $(s,t)$ are as follows:

\begin{small}
\begin{align}
\begin{split}
%\hspace{-6cm}
f_{s,t}^{S} &= \sigma(W_{xf^{S}}*X_{s,t} + W_{h_{s}f^{S}}*H_{s-1,t} + W_{h_{t}f^{S}}*H_{s,t-1} + b_{f^{S}}) \\ 
f_{s,t}^{T} &= \sigma(W_{xf^{T}}*X_{s,t} + W_{h_{s}f^{T}}*H_{s-1,t} + W_{h_{t}f^{T}}*H_{s,t-1} + b_{f^{T}}) \\ 
i_{s,t} &= \sigma(W_{xi}*X_{s,t} + W_{h_{s}i}*H_{s-1,t} + W_{h_{t}i}*H_{s,t-1} + b_{i}) \\ 
\widetilde{C}_{s,t} &= {\rm tanh}(W_{x\widetilde{C}}*X_{s,t} + W_{h_{s}\widetilde{C}}*H_{s-1,t} + W_{h_{t}\widetilde{C}}*H_{s,t-1} + b_{\widetilde{C}}) \\
o_{s,t} &= \sigma(W_{xo}*X_{s,t} + W_{h_{s}o}*H_{s-1,t} + W_{h_{t}o}*H_{s,t-1} + b_{o}) \\
C_{s,t} &= f_{s,t}^{S}\odot C_{s-1,t} + f_{s,t}^{T}\odot C_{s,t-1} + i_{s,t}\odot \widetilde{C}_{s,t} \\
H_{s,t} &= o_{s,t}\odot {\rm tanh}(C_{s,t})
\end{split}
\end{align}
\end{small}
where the input $X_{s,t}$, cell $C_{s,t}$, hidden states $H_{s-1,t}$ and $H_{s,t-1}$, and gates $f_{s,t}^{S}$, $f_{s,t}^{T}$, $i_{s,t}$, $\widetilde{C}_{s,t}$, $o_{s,t}$ are all 3D tensors with dimensions of $M\times N\times F$ (rows, columns, feature maps). \zzz{More precisely, $X_{s,t}$ in Eq. (1) and Eq. (2) represents the feature maps (i.e., $8 \times 8 \times 8$ bottleneck in Fig. \ref{fig_architecture_feature}) after the convolutional encoder on the input image.}

The unit of ST-ConvLSTM (1,1) does not have any preceding units in both the spatial and temporal directions, and units at time 1 level do not have the preceding units in their temporal direction. Zeros activations are fed into these units. The output hidden state of the last unit at time 1 level carries all the tumor information at time 1, thus bringing the global contexts to time 2 through the link connecting itself and the first unit at time 2. It is worth mentioning that the ST-ConvLSTM network is flexible that it can be easily extended to receive more numbers of input time points or to predict longer future steps by recursively applying the model. \zzz{Moreover, \ling{for 3D+time segmentation task,} only sparsely-labeled \ling{manual segmentations} are required, \ling{e.g., representative volumes or even slices.}}

\subsubsection{End-to-End Architecture}

We embed the ST-ConvLSTM unit in the encoder-decoder architecture \cite{finn2016unsupervised,kalchbrenner2017video} to make the end-to-end predictions, as shown in Fig. \ref{fig_architecture_feature} to replace the ST-ConvLSTM unit in Fig. \ref{fig_STCLSTM_cell}. \zzz{In other words, Fig. \ref{fig_architecture_feature} happens in every ST-CLSTM unit in Fig. \ref{fig_STCLSTM_cell}.} Specifically, each frame $X_{s,t}$ in the 4D spatio-temporal space is recurrently passed into the encoder which consists of four convolutional layers to encode a feature map. Along with the image features, clinical factors have non-neglectful influences on predicting the future image as well. We integrate the related factors into our model by spatially tiling the factors (i.e., $m$-dim vector) as a feature map with $m$-channels ($m$=1 in this paper, where only the time interval is added), which is then concatenated to the output of $conv4$ which possesses the smallest number of channels. The concatenated feature map is then fed into a standard ST-ConvLSTM unit (Fig. \ref{fig_STCLSTM_cell}) with a 3$\times$3 kernel and 8 hidden states for the spatio-temporal modeling. As such, the ST-ConvLSTM determines the future state by jointly considering or integrating the compact spatial information of the current slice, the states of slices from previous times and adjacent locations, and clinically relevant factor(s). After that, the decoder with four deconvolutional layers generates the future frame $Y_{s,t+1}$. Because having a smaller transitional kernel helps ConvLSTM to capture smaller motions \cite{xingjian2015convolutional}, we use a 3$\times$3 convolutional \zzz{kernel} by taking into account the knowledge prior that the pancreatic tumor in our dataset is slow-growing. \zzz{For fast-growing tumors, such as glioma, a larger convolutional kernel should be used.} %Additionally, alternative network architectures, such as skip and residual connections \cite{finn2016unsupervised,kalchbrenner2017video} could complement this model. 

\subsubsection{Network Training and Testing}
\zzz{For the tumor growth prediction task}, during training, tumor image slices from time 1 and time 2 are fed as inputs into our network according to their corresponding spatio-temporal locations. Image slices from time 2 and time 3 are used to compute training loss. The objective function of our ST-ConvLSTM network is designed to minimize the $\ell_{2}$ loss between the predicted frames $Y$ and the true future frames $X$ at time 2 and time 3 (other losses, such as $\ell_{1}$ and GDL \cite{mathieu2015deep} have been tried, but $\ell_{2}$ produces empirically better results in our preliminary experiment):
\begin{equation}
L(X,Y) = \sum_{t=1}^{2} \sum_{s=1}^{S} \ell_{2}(Y_{s,t+1},X_{s,t+1})
\end{equation}
where $S$ is the spatial sub-sequence length (set to 5 in our current method). \zzz{Note that minimizing the $\ell_{2}$ loss only at time 3 will have slightly lower performances, and more importantly, cannot maintain a reasonable performance on predicting time 2 which is not desired.} 

In testing, each spatial sequence (at time 1 and time 2) is divided to several sub-sequences, and fed into our model to generate predictions for time 3. These sub-sequences can be either overlapping or non-overlapping. In our preliminary experiment, no significantly differences are ever observed, so we use the non-overlapping sub-sequences for efficiency. In addition, our model is flexible to be extended to make prediction at \zzz{an arbitrary later time point given the observational data of two previous time points}, e.g., predicting time 4 \zzz{based on time 1 and time 2}, by directly setting the value of factor (as depicted in Fig. \ref{fig_architecture_feature}) as the time interval between time 2 and time 4. \zzz{For the problem of 4D ultrasound image sequence segmentation, refer the network training and testing details to Sec. \ref{sec:ultrasound}.}

\section{Experiments}

\subsection{Data and Protocol}

\begin{figure}[!t]
   \begin{center}
   \begin{tabular}{c}
   \includegraphics[width=8cm]{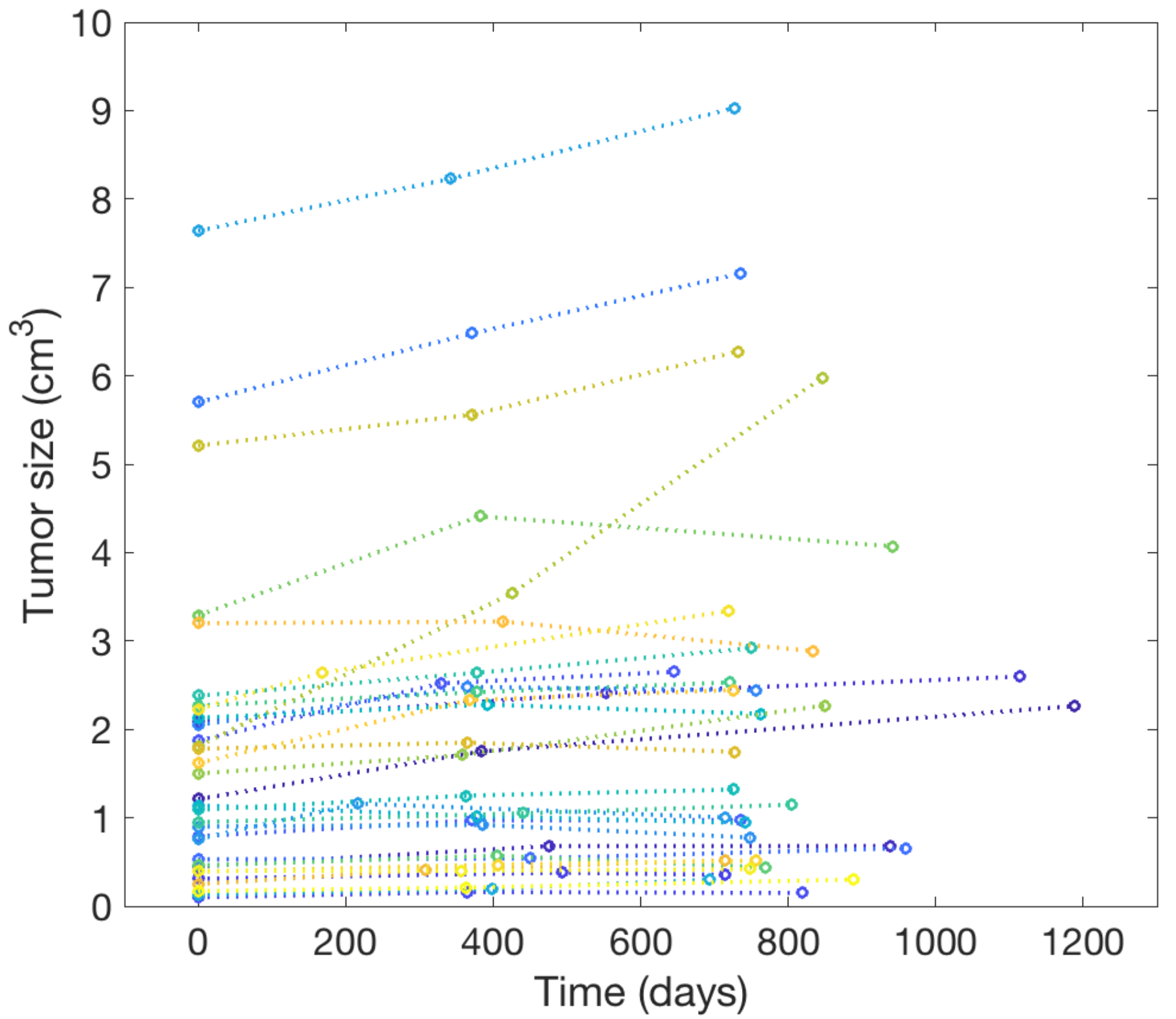}
   \end{tabular}
   \end{center}
   \caption
   { \label{fig_volume_time} 
\zzz{Longitudinal trajectories of PanNET tumor volumes over a population of 33 patients, from time 1 to time 3.}
}
   \end{figure}

Thirty-three patients (thirteen males and twenty females) each with a PanNET are collected from the von Hippel-Lindau (VHL) clinical trial at the National Institutes of Health. Each patient has at least three time points (eleven of these patients have the 4th time points) of dual-phase contrast-enhanced CT imaging, with the time interval of 398$\pm$90 days (average$\pm$std). The CT voxel sizes range between $0.60 \times 0.60 \times 1$ mm$^3$ ---  $0.98 \times 0.98 \times 1$ mm$^3$, \zzz{and are resampled to $1 \times 1 \times 1$ mm$^3$ by trilinear interpolation}. \zzz{We did not include the modality of FDG-PET imaging as it only exists in a small portion of patients or time points. We acknowledge that the prediction performance may not be optimal without PET information. Nevertheless a CT-only predictive model can have wider application scenario (e.g., when more specific PET imaging is not available).} The average age of patients and average volume of tumors at time 1 are 50$\pm$11 years and 1.7$\pm$1.7 cm$^{3}$, respectively. \zzz{Fig. \ref{fig_volume_time} shows the trajectories of PanNET growth rates for different patients.} The average information of all 33 patients is shown in Table \ref{tabletumorinfor}. These tumors keep \zzz{slowly} growing in general, but the growth speed is lower in the 2nd-3rd time period. From the 1st to 2nd time points, only one tumor shrinks. Such a number changes to twelve from the 2nd to 3rd time points. Only 11 patients have real imaging data and time interval information at time 4.

%-------------
%-------------

% --------- Table ------------ %
\begin{table}[!t]
\begin{center}
\tabcolsep=0.11cm
%\newcolumntype{C}{>{\centering\arraybackslash}p{5em}} % For equal spacing below multicolumn
\caption{Statistics of tumor growth at the 1st, 2nd, and 3rd time points of 33 patients.}
\label{tabletumorinfor}
\footnotesize
\begin{tabular}{ p{1.1cm}p{1cm}p{1.4cm}p{1cm}p{1.4cm}p{1.8cm} }
    \toprule
    & \multicolumn{2}{l}{1st-2nd} & \multicolumn{2}{l}{2nd-3rd} &   \\
    \cmidrule(lr){2-3} \cmidrule(lr){4-5}
    {} & {Days} & {Growth (\%)}  & {Days} & {Growth (\%)} & {Size (cm$^{3}$, 3rd)}  \\
    \midrule
    Average & 379$\pm$68 & 24.0$\pm$23.1 & 416$\pm$105 & 8.8$\pm$19.7 & 2.2$\pm$2.2 \\
    {[min,max]} & [168,553] & [-10.5,95.6] & [221,804] & [-23.2,68.8] & [0.1,9.0] \\
    \bottomrule
\end{tabular}
\end{center}
\end{table}
% ---------------------------- %

\subsection{\zzz{Experimental Design \&} Implementation Details}

{\bf Training details:} Four data augmentation schemes are performed to enrich our dataset. Besides the original axial image slice sequences, we 1) reformat\zzz{/reslice original volumes to obtain} coronal and sagittal slices, 2) rotate (with 90 degree interval), 3) translate\zzz{/shift} (randomly 2 pixels in $xy$ plane) for each 4D ICVF-CT-Mask volumetric sequence, and 4) reverse the \zzz{slice order in spatial direction}. \zzz{Then, $S$=5 sub-sequences are cropped from the augmented sequences \ling{(the minimal-sized tumor in our dataset has about 5 slices)}, resulting} in 172,296 training sub-sequences in total. Such methods add more variations into the generated or augmented dataset and improve the generalization capability. \zzz{Note that we ensure the augmented sequences are still spatio-temporally aligned.} We train our ST-ConvLSTM models for 5 epochs with the batch size of 16. Each data point has 5 slices at three time points. We use the ADAM optimizer \cite{Kingma2015Adam} for neural network optimization with an initial learning rate of 10$^{-3}$.  

\zzz{{\bf Testing details:} In the testing scenario, given any testing PanNET data including pre- and post-contrast CT scans from time 1 and time2 ($X_{t} = \{X^{i}_{t},X^{c}_{t},X^{m}_{t}\}, t\in\{1,2\}$), the preprocessing steps in Sec. \ref{dataconstruction} are applied to first obtain the aligned spatio-temporal ICVF-CT-Mask sequence pair. Next we divide the aligned data into \ling{several} non-overlapping sub-sequence pairs: each sub-sequence image pair containing 5 consecutive slices from time 1 and their corresponding 5 from time 2. By feeding \ling{these} sub-sequence pairs together with the time interval features as (time1--time2) and (time2--time3) into our model, the future (at time 3) consecutive data slices ($Y_{t} = \{Y^{i}_{t},Y^{c}_{t},Y^{m}_{t}\}, t=3$) can be predicted and produced. A thresholding value of 128 is applied upon the predicted probability map of mask channel to obtain a binary tumor mask $Y^m$.}

{\bf Comparison:} We implement the current clinical practice of a default linear growth model, the conventional ConvLSTM \cite{xingjian2015convolutional}, and another major deep learning method for video prediction, i.e., BeyondMSE (GAN) \cite{mathieu2015deep}, for model comparison. The linear growth model assumes that tumors would keep their past growing trend in the future. As detailed in \cite{zhang2018convolutional}, the past radial expansion/shrink distances on tumor boundaries are used to expand/shrink the current tumor boundary as future prediction. The ConvLSTM uses the same architecture as in Fig. \ref{fig_architecture_feature} (but it only captures the temporal dependencies) and is trained with the same network optimization setting as ST-ConvLSTM. In the BeyondMSE framework, a multi-scale fully convolutional ConvNet is used as the future image generator, and a multi-scale ConvNet as the discriminator. The generator receives two past images as input and outputs one future image, while the discriminator receives all three images as input to classify whether they are real or fake. Our implementation uses the same network architecture and parameter setting as in \cite{mathieu2015deep}. Both ConvLSTM and BeyondMSE are trained for 5 epochs on the same augmented dataset as ST-ConvLSTM. All these aforementioned models are implemented in TensorFlow \cite{tensorflow2015} and perform experiments on a DELL TOWER 7910 workstation with 2.40 GHz Xeon E5-2620 v3 CPU, 32 GB RAM, and a Nvidia TITAN X Pascal GPU of 12 GB of memory. Note that compared to previous machine (deep) learning based tumor growth model prediction methods \cite{morris2006learning,zhang2017personalized,zhang2018convolutional} that merely utilize 2D image patches and only predict the future tumor volume, our new ST-ConvLSTM model is holistically 4D (volumetric+time) image-based and enables the predictions of future tumor imaging properties, such as future cell density and CT intensity numbers to assist relevant clinical diagnosis.

\zzz{{\bf Predicting a later future:} In this experiment, we evaluate the problem of predicting a later time step 4 given only time 1 and time 2 available. For those 11 patients who have follow-up studies at time 4, we directly set the time interval between time 2 and time 4 as the feature in the trained predictive model. For the remaining 22 patients without the follow-up time step 4, we assume that their time 3 and time 4 have the equal time interval as the interval between their time 2 and time 3, in order to investigate the effectiveness of time interval feature in our predictive model on a larger patient cohort (of all 33 patient data).
}

\subsection{Evaluation Methods}
We evaluate our model using three-fold cross-validation. In each fold, 22 patients are used as training and the remaining 11 patients as testing data. The performance of tumor prediction is evaluated at the 3$^{rd}$ time point by the metrics of Dice coefficient and RVD (relative volume difference) \cite{zhang2017personalized,zhang2018convolutional,wong2017pancreatic} for tumor volume, RMSE (root-mean-squared error) for ICVF \cite{wong2017pancreatic}, and diff.HU (difference of average HU values) for CT value. \begin{align}\label{metrics}
\begin{split}
&{\rm Dice} = \frac{2 \times \rm TPV}{V_{pred} + V_{gt}} \\ 
&{\rm RVD} = \frac{|V_{pred} - V_{gt}|}{V_{gt}} \\
&{\rm RMSE} = \sqrt{ \frac{\sum(( icvf_{pred}-  icvf_{gt})/  icvf_{gt})^{2}}{\rm TPV} } \\
&{\rm diff. HU} = \frac{|\rm HU_{pred}- \rm HU_{gt}|}{\rm HU_{gt}}
\end{split}
\end{align}
where TPV (true positive volume) is the overlapping volume between the predicted tumor volume $V_{pred}$ and the ground truth tumor volume $V_{gt}$. $icvf$ represents the ICVF value of a pixel. HU represents the average Hounsfield units within a volume. Both RMSE and diff.HU are evaluated within the \ling{TPV following \cite{wong2017pancreatic}, in which RMSE of ICVF prediction is assessed in the TPV}. Paired $t$-tests are conducted to compare our new model and other previous methods.

\zzz{We calculate the scatter plots of the ST-ConvLSTM predicted tumor volumes and the respective growth rates (Fig. \ref{fig_scatterplot}), in comparison with the ground truth. Based on the tumor growth rate, we also assess the performance of our method on another clinical relevant prediction task -- prediction of tumor progression vs. regression at time 3. Specifically, the prediction results are divided in two groups comprising tumor progression (positive growth rate) and tumor regression (negative growth rate), where sensitivity and specificity are used as evaluation metrics. As an alternative solution, traditional machine learning methods are applied on this task by training binary classifiers and evaluating using the same three-fold cross-validation. Specifically, age, gender and tumor volume measures at time 1 and time 2, tumor volume changes between time 1 and time 2, are extracted as features for classification. Optimal feature combinations are experimented and assessed by several common classifiers including logistical regression, linear SVM, neural network (one hidden layer), and random forest.} 

\zzz{We examine the contribution of each input feature channel, by training with only one and predicting the corresponding future one: for example, from given previous CT scans to predict a future CT image. Note that RMSE and diff.HU reported here are also evaluated within the true positive volume.}

\subsection{Quantitative Results}

%-------------
   \begin{figure*}[!t]
   \begin{center}
   \begin{tabular}{c}
   \includegraphics[width=18cm]{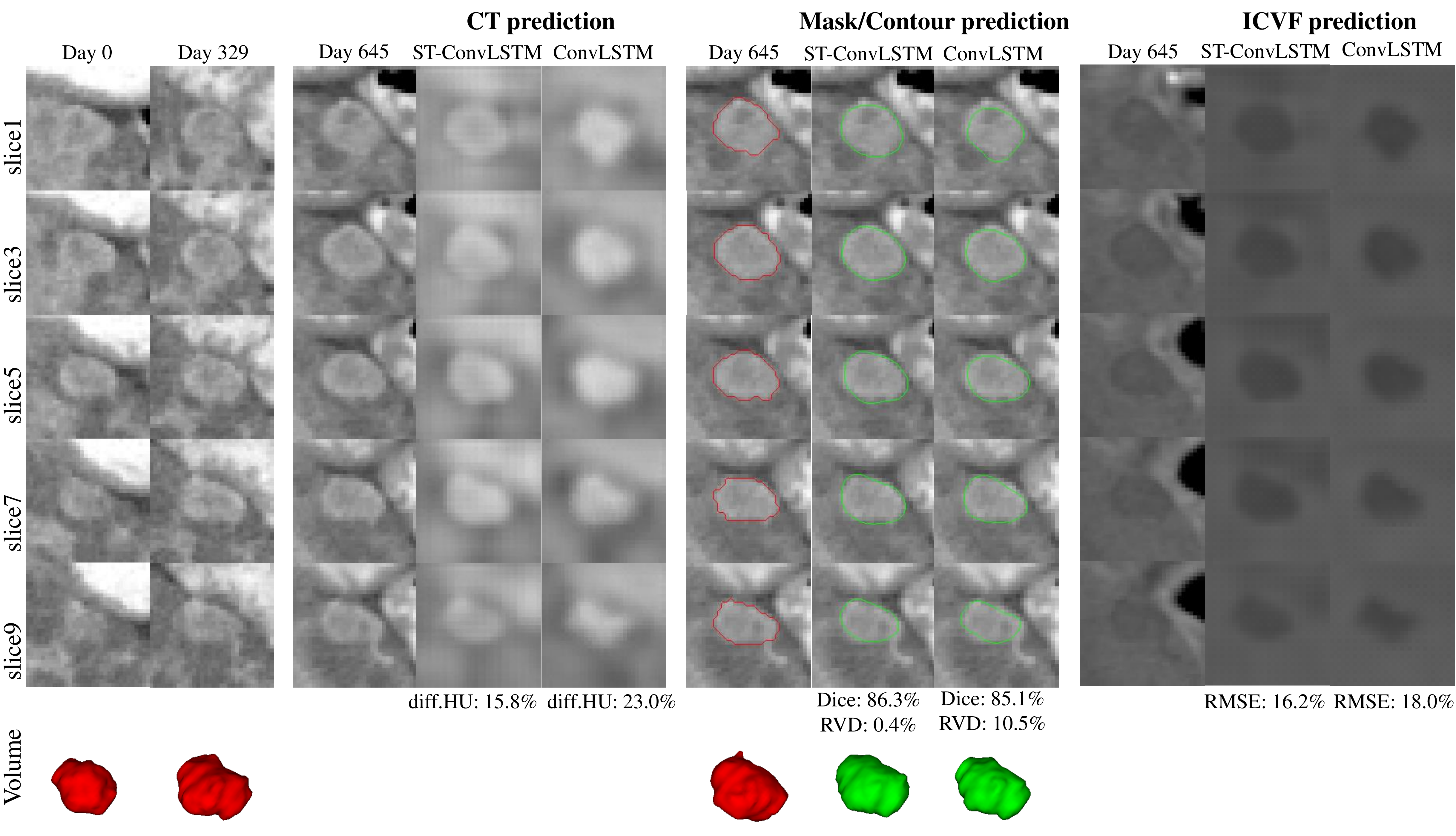}
   \end{tabular}
   \end{center}
   \caption
   { \label{fig_result} 
An illustrated example shows the prediction results of CT, mask/volume, and ICVF of a tumor by ST-ConvLSTM and ConvLSTM. Note that the tumor contours are superimposed on the ground truth CT images at time 3. Red: ground truth boundaries; Green: predicted tumor boundaries. In this example, consecutive image slices with the spatial interval of two slices are shown for better visualization of the spatial changes/differences.
}
   \end{figure*} 
%-------------

\begin{table*}[!t] 
\scriptsize
\centering
\caption{Overall quantitative performance on 33 patients under 3-fold cross-validation -- Baseline linear predictive model, ConvLSTM \cite{xingjian2015convolutional}, and our ST-ConvLSTM. Results are reported as: mean $\pm$ std [min, max]. \zzz{* or ** indicates a statistically significant difference of our method compared to other methods and BeyondMSE (GAN), respectively.} \zzz{There is no significant difference between ST-ConvLSTM with and without time interval feature.}
}
\label{performance}
\begin{tabular}{p{2.7cm}|p{2.65cm}|p{2.65cm}|p{2.4cm}|p{2.4cm}}
\hline
 & Volume-Dice (\%) & Volume-RVD (\%) & ICVF-RMSE (\%) & CT-HUdiff. (\%)\\
\hline
Linear & 73.0$\pm$6.2 [60.2, 85.1] & 22.8$\pm$18.3 [5.1, 75.2] & - & - \\
\hline
ConvLSTM \cite{xingjian2015convolutional} & 82.1$\pm$5.8 [65.6, 90.4] & 14.1$\pm$12.4 [1.2, 50.4] & 13.7$\pm$8.4 [6.8, 42.4] & 10.4$\pm$8.3 [0.6, 32.4] \\
\hline
BeyondMSE (GAN) \cite{mathieu2015deep} & 79.3$\pm$5.7 [65.6, 90.4] & 20.9$\pm$14.4 [1.2, 50.4] & 19.7$\pm$12.0 [6.8, 42.4] & 10.7$\pm$8.1 [0.6, 32.4] \\
\hline
\zzz{ST-ConvLSTM w/o. time} & \zzz{83.1$\pm$4.9* [67.9, 91.1]} & \zzz{12.6$\pm$9.0* [0.3, 48.6]} & \zzz{13.9$\pm$7.9** [7.8, 43.5]} & \zzz{10.0$\pm$7.3 [0.2, 26.3]} \\
\hline
ST-ConvLSTM \zzz{w. time} & 83.2$\pm$5.1* [69.7, 91.1] & 11.2$\pm$10.8* [0.3, 46.5] & 14.0$\pm$8.5** [7.4, 41.4] & 10.2$\pm$8.5 [0.0, 35.0] \\
\hline
\end{tabular}
\end{table*} 

The visual example in Fig. \ref{fig_result} shows the prediction results of future CT scan, tumor mask/volume, and ICVF obtained by ST-ConvLSTM \zzz{(with time interval feature)} and ConvLSTM. In this case, compared with the conventional ConvLSTM, our ST-ConvLSTM generates more spatially consistent prediction \zzz{towards the actual tumor in terms of} CT, mask and ICVF, and consequently, achieves better accuracies under all quantitative metrics \zzz{(i.e., diff.HU, Dice, RVD and RMSE)}. Table \ref{performance} reports the overall performance of our ST-ConvLSTM model \zzz{(with and without time interval feature)} with that of ConvLSTM and the linear model on 33 patients. For the volume prediction, ST-ConvLSTM \zzz{(w. time)} produces a Dice score of 83.2\% and a RVD of 11.2\%. Both are significantly better than ConvLSTM ($p < $0.01 and $p < $0.05) and linear predictive model ($p < $0.001 and $p < $0.01). Furthermore, our model generates a RMSE of 14.0\% for tumor cell density prediction, and a diff.HU  of 10.2\% for radiodensity prediction (no statistical significances are achieved on these two metrics in comparison to ConvLSTM). \zzz{There is no significant difference between ST-ConvLSTM with and without time interval feature.}

\begin{figure*}[!t]
   \begin{center}
   \begin{tabular}{c}
   \includegraphics[width=17.5cm]{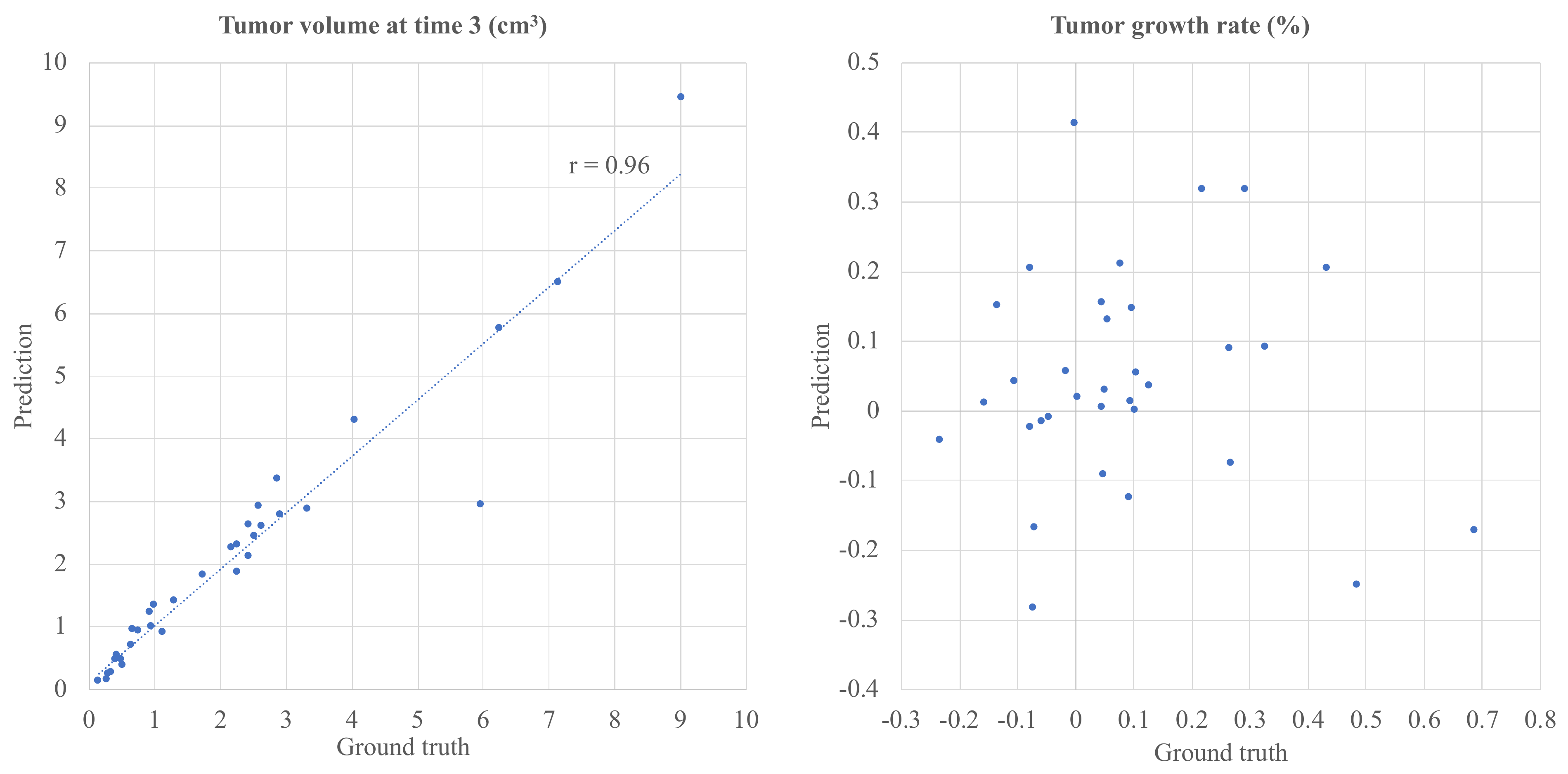}
   \end{tabular}
   \end{center}
   \caption
   { \label{fig_scatterplot} 
\zzz{Scatter plots of ST-ConvLSTM predicted tumor volumes versus ground truth values \ling{($r$ is the linear correlation coefficient)} (\textbf{Left}) and predicted tumor growth rates versus true rates (\textbf{Right}).}
}
   \end{figure*}

\zzz{The ST-ConvLSTM predicted tumor volumes achieve high correlations against the ground truth volumes (\ling{linear correlation coefficient} $r$=0.96, the left panel in Fig. \ref{fig_scatterplot}). However, the prediction of tumor growth rate is highly challenging (with $r$=0.04, the right panel in Fig. \ref{fig_scatterplot}), especially for some extreme cases such as tumor shrink and aggressive progression (quadrants II and IV in the right panel in Fig. \ref{fig_scatterplot}). Thus we assess the performance of a relatively more convenient binary prediction task: tumor progression (positive growth rate, 21 patients) versus regression (negative growth rate, 12 patients). As shown in Table \ref{progvsreg}, our method has a sensitivity of 76.2\% and specificity of 50\%, in compared to an optimized machine learning approach on this dataset that achieves a sensitivity of 61.9\% and specificity of 50\%. This result is obtained by a random forest classifier using the tumor volume at time 1 as the only feature (other feature combinations and classifiers are empirically worse than this performance). The overall available features and patient numbers for our studied problem are still limited.}

%-------------
%-------------

\begin{table}[!t]
\scriptsize
\centering
\caption{\zzz{Performance of predicting tumor progression versus regression by ST-ConvLSTM and an optimized machine learning approach on this dataset.}}
\label{progvsreg}
\begin{tabular}{l|c|c}
\hline

                 & \multicolumn{1}{l|}{\zzz{Sensitivity (\%)}} & \multicolumn{1}{l}{\zzz{Specificity (\%)}} \\ \hline
\zzz{Time 1's tumor volume + random forest} & \zzz{61.9}                                  & \zzz{50.0}                                 \\ \hline
\zzz{ST-ConvLSTM}      & \zzz{76.2}                                  & \zzz{50.0}                                 \\ \hline

\end{tabular}
\end{table}

\begin{figure*}[!t]
   \begin{center}
   \begin{tabular}{c}
   \includegraphics[width=18cm]{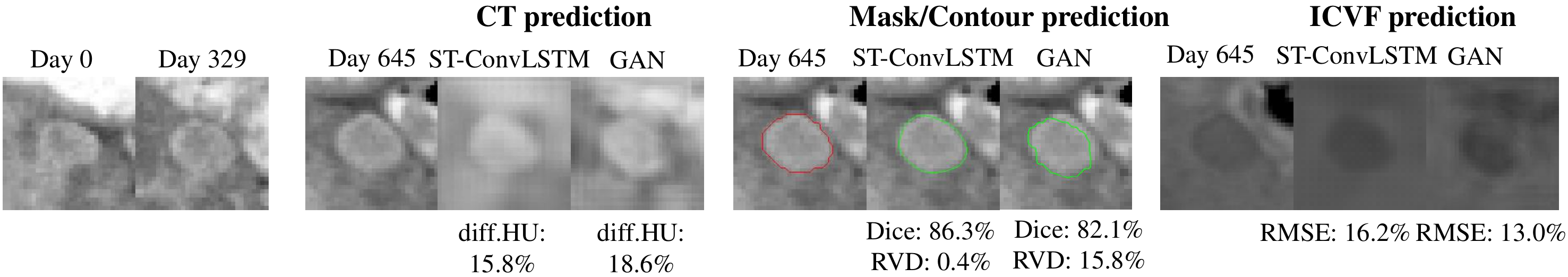}
   \end{tabular}
   \end{center}
   \caption
   { \label{fig_gan} 
An example of image slices shows the prediction results of CT, mask/volume, and ICVF of a tumor by ST-ConvLSTM and BeyondMSE (GAN). Note that the tumor contours are superimposed on the ground truth CT images at time 3. Red: ground truth boundaries; Green: predicted tumor boundaries.
}
   \end{figure*}

Figure \ref{fig_gan} compares the prediction results of our ST-ConvLSTM with BeyondMSE (GAN) ~\cite{mathieu2015deep}. In this case, BeyondMSE has reported noticeably worse performance in predicting tumor volume, but generates less blurry CT and ICVF images (through visually observation). Table \ref{performance} lists the overall prediction performance of BeyondMSE, where the proposed method significantly outperforms BeyondMSE in terms of Dice, RVD, and ICVF-RMSE.

%-------------
%-------------

\begin{figure*}[!t]
   \begin{center}
   \begin{tabular}{c}
   \includegraphics[width=17.5cm]{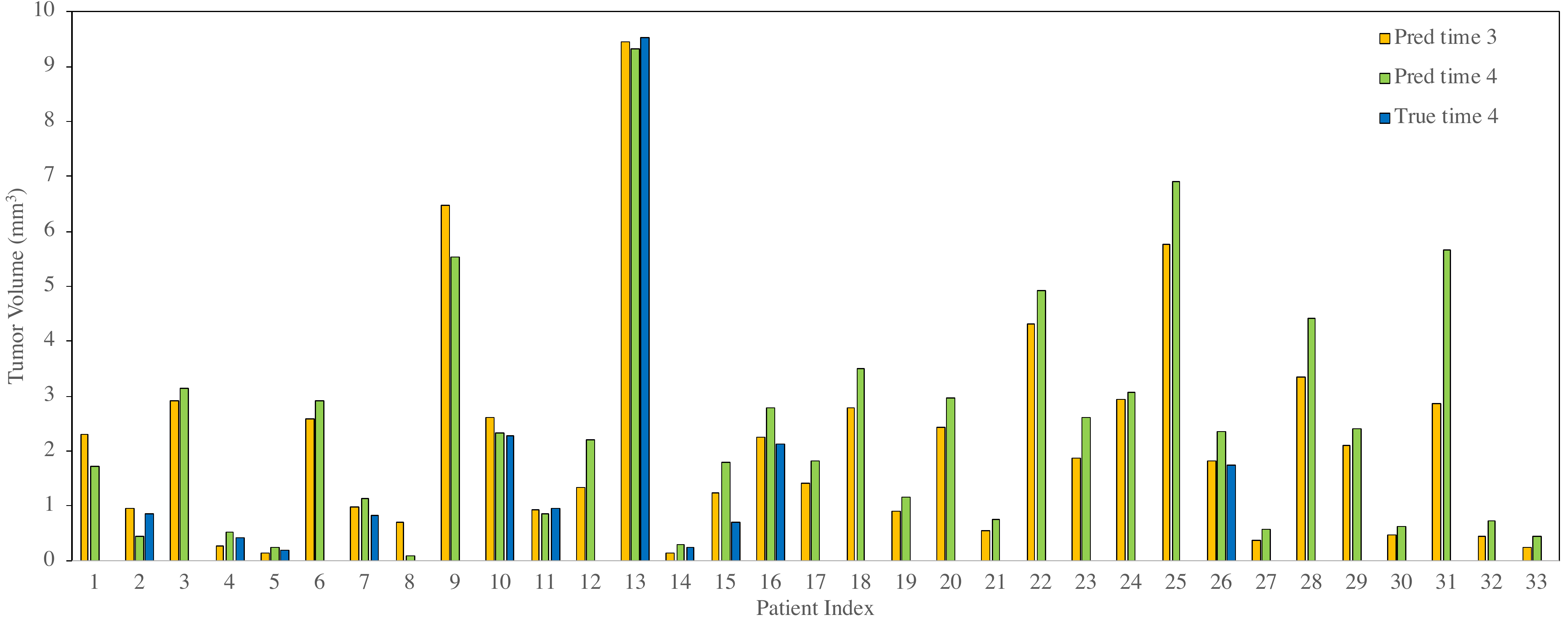}
   \end{tabular}
   \end{center}
   \caption
   { \label{fig_laterfuture} 
ST-ConvLSTM prediction results at an even later time point. Volume prediction results at time 4 based on time 1 and time 2 for all 33 patients. 26 out of 33 (78.8\%) patients are predicted as tumor keeping growing (i.e., tumor size at time 4 larger than time 3). %{\bf Lower panel}: CT, mask, and ICVF predictions at time 4 for a slice from case 6. Note that time 3's results are also shown for reference; the tumor contours are superimposed on the predicted CT images at time 4 (since the ground truth CT images at time 4 is not available for this patient). Red: ground truth boundaries; Green: predicted tumor boundaries.
}
   \end{figure*}

\begin{figure}[!t]
   \begin{center}
   \begin{tabular}{c}
   \includegraphics[width=8.5cm]{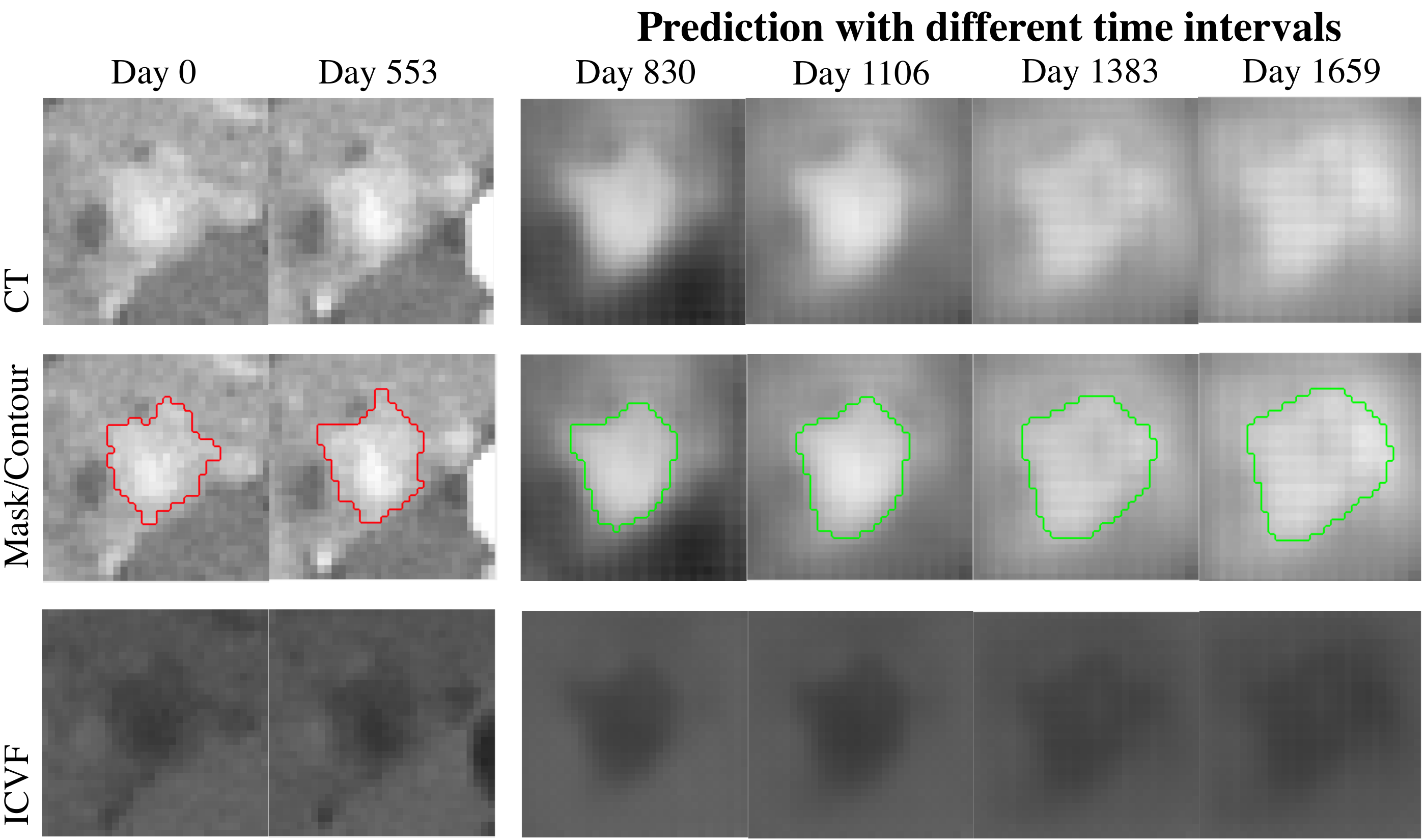}
   \end{tabular}
   \end{center}
   \caption
   { \label{fig_laterfuture_R1} 
\zzz{One tumor example using ST-ConvLSTM predictions from time 1 (Day 0) and time 2 (Day 553), and at different later time points (Day 830, 1106, 1383, 1659). Note that the predicted tumor becomes larger when time interval grows.}
}
   \end{figure}

Fig. \ref{fig_laterfuture} shows the prediction results at an even later time step using ST-ConvLSTM for all 33 patients. As a result, 78.8\% tumors are predicted to keep growing at later time points -- the predicted volume at time 4 is larger than time 3. For the 11 tumors which have ground truth measures of tumor volume at time 4, our prediction produces a RVD of 37.2\%$\pm$42.5\%. \zzz{Fig. \ref{fig_laterfuture_R1} illustrates qualitative tumor visualization results upon changing the time interval feature, to examine the model's predictions at future possible time steps.} \zzz{From Table \ref{ablation}, when using the tumor mask as the single input channel, it produces statistically similar Dice and RVD measures as three input channels are utilized. However the three-channel ICVF-CT-Mask input configuration generates clearly better performance on RMSE and HUdiff. predictions.}

%-------------
%-------------

%-------------
%-------------

\begin{table}[!t]
\scriptsize
\centering
\caption{\zzz{Ablation study showing results for different input feature channels. * indicates a statistically significant difference.}}
\label{ablation}
\begin{tabular}{l|c|c|c|c}
\hline
\zzz{Input channel} & \zzz{Dice (\%)}    & \zzz{RVD (\%)}      & \zzz{RMSE (\%)}     & \zzz{HUdiff. (\%)}  \\ \hline
\zzz{ICVF}          & \zzz{-}            & \zzz{-}             & \zzz{15.7$\pm$8.9} & \zzz{-}             \\ \hline
\zzz{CT}            & \zzz{-}            & \zzz{- }            & \zzz{-}             & \zzz{12.1$\pm$10.8} \\ \hline
\zzz{Mask}          & \zzz{83.6$\pm$4.7} & \zzz{13.7$\pm$11.9} & \zzz{-}             & \zzz{-}             \\ \hline
\zzz{ICVF+CT+Mask}           & \zzz{83.1$\pm$4.9} & \zzz{12.6$\pm$9.0} & \zzz{13.9$\pm$7.9*}  & \zzz{10.0$\pm$7.3*}  \\ \hline
\end{tabular}
\end{table}

On average, our method takes $\sim1.2$ hrs for training and 0.2 second for prediction per tumor. This performance is faster than the statistical and deep learning framework ($\sim$ 3.5 hrs training and 4.8 mins prediction \cite{zhang2017personalized}) in both training and inference; while faster than the two-stream ConvNets \cite{zhang2018convolutional} in prediction but slower in training.

\subsection{\zzz{Segmentation in 3D+Time Ultrasound}}\label{sec:ultrasound}
\zzz{
To further demonstrate the feasibility of ST-ConvLSTM for 4D segmentation, the publicly available 3D+time ultrasound dataset CETUS \cite{bernard2015standardized} is used. The dataset is acquired from 15 patients where each patient containing 13--46 3D volumetric image sequences and each sequence with two volumes being manually segmented at the end-diastole (ED) and endsystole (ES) phases. We resample all 3D ultrasound scans to 1mm$^{3}$ isotropic resolution. For facilitating ST-ConvLSTM training, the 4th dimension is downsampled to a constant length (i.e., 6 time points in our work), with image annotation/segmentation masks at the 1st and 6th time points. All image slices are resized to 96$\times$96 pixels and pixel intensities are normalized to [0, 1]. 
}

\zzz{
2D ultrasound image slices are fed as inputs into the network according to their corresponding spatio-temporal locations to generate the corresponding segmentation masks, via a $\ell_{2}$ training loss computing only at two \ling{volumes from two} time points (ED and ES phases). We train our model for 30 epochs with the batch size of 1. Each data instance has 10 image slices at 6 time points. We use ADAM optimizer \cite{Kingma2015Adam} for the network optimization with an initial learning rate of 10$^{-3}$. In testing, each sequence is divided to several sub-sequences and fed into our model to generate the segmentation mask at each time. The segmentation masks are post-processed with the largest connected-component selection. Five-fold cross-validation at the patient-level splitting is conducted.
}

\begin{figure*}[!t]
   \begin{center}
   \vspace{-3pt}
   \begin{tabular}{c}
   \includegraphics[width=17cm]{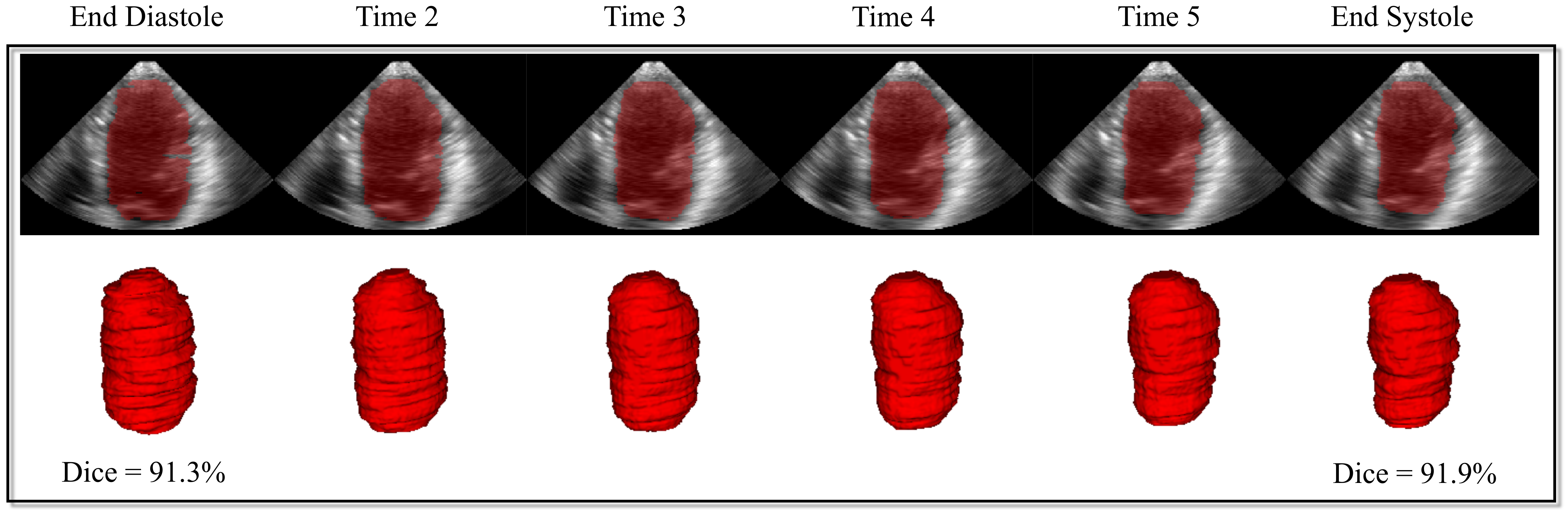}
   \end{tabular}
   \vspace{-12pt}
   \end{center}
   \setlength{\belowcaptionskip}{-5pt}
   \caption
   { \label{fig_4dseg} 
\zzz{4D segmentation results of left ventricle in 3D+time ultrasound by ST-ConvLSTM.}
}
   \end{figure*}

\zzz{
Our method achieves the segmentation performance of Dice at 86.8\%$\pm$2.1\% and 85.9\%$\pm$1.6\% for ED and ES phase, respectively. Compared to the CETUS 2014 challenge winner (89.4\%$\pm$4.1\% and 85.6\%$\pm$5.7\%, using deformable model approach) \cite{bernard2015standardized}, our method performs better for ES but worse for ED. Fig. \ref{fig_4dseg} shows an example of our 4D segmentation result from ED to ES. Our method is efficient by taking $\sim6.5$ hrs for training of 30 epochs and only 3 seconds for segmentation in testing per 4D sequence (6 ultrasound volumes in our setting). 
}

%-------------
%-------------

\subsection{Discussion}
Deep learning based precision and predictive medicine is a new emerging research area, and has been shown to be capable of outperforming traditional mathematical modeling based methods for tumor growth prediction. This may suggest its great potential for solving this complicated but important problem. Because of the tremendous difficulties of collecting the longitudinal tumor data, most previous studies are evaluated on a relatively small sized dataset (i.e., $<$ 10 patients). A statistically larger and more representative patient dataset is desired to evaluate the prediction performance. Our novel model, ST-ConvLSTM network, significantly differs from the most recent statistical and deep learning \cite{zhang2017personalized} and two-stream ConvNets \cite{zhang2018convolutional} in several key aspects. Firstly, it uses a single recurrent neural network to explicitly and jointly model the temporal changes and spatial consistency (i.e., in 4D space), rather than separate invasion and expansion networks to model the temporal information only (i.e., 2D+time) \cite{zhang2017personalized,zhang2018convolutional}. Secondly, it makes prediction at the holistic image-level instead of local image patch-level, integrating the global spatial context of tumor structure and meanwhile being more computationally efficient. Thirdly, it enables the prediction of both future images and the associated imaging properties, including CT scan, tumor cell density and radiodensity, as demonstrated in this paper. Fourthly, it uses an encoder-decoder deep neural network architecture that incorporates imaging feature and clinical factor (such as time interval) in an end-to-end learning framework, rather than a late feature fusion stage. Fifthly, we \zzz{construct} the largest longitudinal tumor dataset (33 patients) to date to the best of our knowledge, and comprehensive quantitative evaluation results against three other prediction methods using ConvLSTM \cite{xingjian2015convolutional} and GAN \cite{mathieu2015deep}. Finally, we extend our deep learning based method to make it capable of predicting any time point in a later future (beyond time point 3).

One of our main contributions is the novelty of proposed ST-ConvLSTM architecture. Compared to the previous state-of-the-art ConvLSTM \cite{xingjian2015convolutional} model for temporal modeling of 2D image sequences across different time points, we substantially extend the ConvLSTM into the spatio-temporal 4-dimensional space by jointly leaning both the temporal evolution of tumor growth and the spatial information of 3D consistency. Particularly, for the adjacent 2D CT slices, they are also modeled by ConvLSTM (slice-by-slice) to ensure their spatial consistency. In addition, the global contexts of previous time point are fed to the current time point. Therefore, each ST-ConvLSTM unit makes prediction not only based on its local spatial and temporal neighbors, but also from the whole information of past states. As a result, our ST-ConvLSTM is able to generate a sequence of images with better 4D properties than ConvLSTM, e.g., producing statistically higher accuracy in volume prediction, as shown in Table \ref{performance}. An illustrative example can be observed from Fig. \ref{fig_result}. ST-ConvLSTM generates more consistent tumor morphology and structure for CT, mask, and ICVF predictions than ConvLSTM results (of irregular predictions for tumor morphology). An alternative option of using ConvLSTM for the 4D prediction task can simply stack 2D CT slices as different input channels and modeling the temporal relation using LSTM. However such a method cannot exploit either the inherent correlations of inter-slice correlations in 3D contexts, or temporal dynamics across time points. The simple linear predictive model performs the worst among all compared methods. This is in agreement with the fact that the pancreatic neuroendocrine tumors demonstrate nonlinear growth patterns \cite{weisbrod2014assessment,keutgen2016evaluation}.  \zzz{The ablation study shows that directly predicting the future tumor mask based on previous masks may perform comparably with the configuration of using all three information (ICVF, CT, mask). This is in accordance with the finding of a computer vision study \cite{luc2017predicting} that segmentation-to-segmentation prediction generates no worse result than RGB+segmentation-to-segmentation prediction. Of course, the complete ICVF-CT-mask configuration offers better performance on RMSE and HUdiff. predictions, and more importantly, can compute the future tumor CT images (Fig. \ref{fig_laterfuture_R1}).}

\zzz{Beyond the tumor prediction task, our proposed novel ST-ConvLSTM architecture can be adapted conveniently for learning 4D medical image representations. We demonstrate its promising accuracy and high efficiency (e.g., 3 seconds to process per 4D imaging sequence) in 4D ultrasound image segmentation, while only requiring sparse image annotation masks (e.g., 2 out of 6 volumes per 4D sequence in our experiment) for training. Furthermore, it can also be applied to 4D classification task by changing the network output.}

Besides ConvLSTM, BeyondMSE (GAN) \cite{mathieu2015deep} is another deep learning model for future frame prediction. Benefited from the $\ell_{1}$, image gradient based optimization and adversarial losses, GAN could generate less blurry future image predictions, as shown in Fig. \ref{fig_gan}. However GAN has much lower quantitative prediction performance than our method. One reason may be that GAN does not explicitly model the temporal dynamics, while LSTM has inherent temporal ``memory'' units though GAN-based tumor prediction can somewhat capture the tumor growing trend. For example, in Fig. \ref{fig_gan}, from time 1 to time 2, the tumor invasion happens mostly in its lower part so that GAN predicts the tumor to continue infiltrating to the below area at time 3. Nevertheless the tumor actually slows down its growing speed at time 3 in that direction. Our ST-ConvLSTM model learns the spatio-temporal information jointly and can leverage the current slice's global and local neighbors' information, which results in more robust prediction. On the other hand, the GAN-based method may have higher overfitting risk on our task. The network architectures used in \cite{mathieu2015deep} can be over-complicated for the relatively small-sized data studied in this work.

\ling{Using time-interval feature in the ST-ConvLSTM does not improve the prediction accuracy compared to without using such feature. This may be because for time 1-2 and time 2-3 in our data, 1) the time-intervals are similar (about 1 year) for different patients, and 2) the PanNet is slow-growing and can show different growing trend. Actually, recent studies show that time feature can be either helpful \cite{baytas2017patient} or not helpful \cite{li2018time} in LSTM-based prediction on different medical data. More investigation is needed in this direction.}
\ling{Nevertheless, the time-interval feature is necessary for tumor growth prediction problem. For example, }for the prospect of longer future prediction of tumor growth, the time-interval feature is effective to control our predictive model to generate sensible prediction results, \zzz{as shown by the illustrative example in Fig. \ref{fig_laterfuture_R1}. Furthermore,} our model predicts that 78.8\% tumors keep growing at time 4. This is in accordance with the natural history of PanNET tumors, around 20\% decreasing over a median follow-up duration of 4 years \cite{weisbrod2014assessment}. \ling{However, considering the missing of ground truths of 22 patients at time 4, the related results and discussions should be treated with caution. For the 11 patients with ground truths at time 4,} the prediction accuracy at a longer future time point (i.e., RVD=37.2\% at time 4) is much lower than that of the next predictable time step (i.e., RVD=\ling{15.7}\% at time 3). This is as expected since it is indeed harder to precisely predict the tumor growth trends and patterns after a longer period of time, for example, around two years later using our data. As a reference, a recent mathematical modeling based tumor growth prediction method \cite{roque2018dce} has the relative volume errors of later time predictions, ranging from 45\% to 123\% for breast carcinoma. Another solution for predicting further into the future is to recursively apply the two-time-input model as in \cite{luc2017predicting}, i.e., predicting the outcome of time 4 based on the time 2 and the predicted time 3 results.  

There are some future directions which may further improve our method. First, the $\ell_{2}$ loss function used in our model is the major reason that causes blurry predictions. Adversarial training \cite{mathieu2015deep} can increase the sharpness of the predicted image and is straightforward to be incorporated into our ST-ConvLSTM network, through using a discriminator to determine whether the generated future image sequence is real or fake during training. Second, alternative network architectures, such as skip and residual connections \cite{finn2016unsupervised,kalchbrenner2017video} may complement our current encoder-decoder network as the backbone. \zzz{Third, testing time data augmentation may further improve the prediction performance, e.g., averaging prediction results along three reconstruction directions: axial, coronal, and sagittal. Fourth, predicting the tumor growth rate is challenge for the current model. This may be caused by the limited training data in which most tumors are slow-growing whereas our model is not trained to directly predict the tumor growth rate. Our deep model can scale well and perform better by incorporating more patient data when available in the future. Another potential solution is explicitly personalizing the predictive model as in our previous work \cite{zhang2018convolutional}. Although obtaining much better result in predicting aggressive progression, we find that it decreases the overall volume prediction accuracy (increasing RVD difference from 11.2\% to 13.1\%) on our dataset. Nevertheless, our model shows competitive results on predicting tumor progression versus regression compared to traditional machine learning approaches.}

\section{Conclusion}
In this paper, we have employed and substantially extended ConvLSTM \cite{xingjian2015convolutional} in the 4-dimensional spatio-temporal domain for the task of modeling 4D longitudinal tumor data. The novel ST-ConvLSTM network jointly learns the intra-slice structures, the inter-slice 3D contexts, and the temporal dynamics. Quantitative results of notably higher accuracies than the original ConvLSTM \cite{xingjian2015convolutional} are reported, using several metrics on predicting the future tumor volumes. Compared to the most recent 2D+time deep learning based tumor growth prediction models \cite{zhang2017personalized,zhang2018convolutional}, our new approach directly works on 4D imaging space and incorporates clinical factors in an end-to-end trainable manner. This method can also predict the tumor cell density and radiodensity. %We also demonstrate that our model has the potential to predict tumor growth in later future. 
Our experiments are conducted on the largest longitudinal pancreatic tumor dataset (33 patients) to date and demonstrate the validity of our proposed method. \zzz{In addition, ST-ConvLSTM enables efficient and effective 4D medical image segmentation with only sparse manual image annotations required.} The presented ST-ConvLSTM model can potentially enable other applications of 4D medical imaging applications. %Future work is to further improve the prediction qualities, likely by involving adversarial training \cite{mathieu2015deep}.

%\section*{Acknowledgments}
%This work was supported by the Intramural Research Program at the NIH Clinical Center.

% Can use something like this to put references on a page
% by themselves when using endfloat and the captionsoff option.
\ifCLASSOPTIONcaptionsoff
  \newpage
\fi

{\small\small
\bibliographystyle{IEEEtran}
\bibliography{LingRef}
}

% that's all folks
\end{document}